\documentclass[10pt,twocolumn,letterpaper]{article}
\usepackage{times}
\usepackage{epsfig}
\usepackage{graphicx}
\usepackage{amsmath}
\usepackage{amssymb}
\usepackage{subcaption} 
\DeclareMathAlphabet{\pazocal}{OMS}{zplm}{m}{n} 
\usepackage{booktabs}
\usepackage{sidecap}
\usepackage{rotating}
\usepackage[breaklinks=true,bookmarks=false]{hyperref}
\usepackage{tikz}

\newcommand\etal[0]{\emph{et al}.}
\newcommand\eg[0]{\emph{e.g.}}
\newcommand\ie[0]{\emph{i.e.}}

\begin{document}

\title{Hue-Net: Intensity-based Image-to-Image Translation with Differentiable Histogram Loss Functions}
\author{Mor Avi-Aharon, Assaf Arbelle, and Tammy Riklin~Raviv \\The Department of Electrical and Computer Engineering\\ Ben-Gurion University of the Negev}
\date{}
\maketitle

\begin{abstract}
We present the Hue-Net - a novel Deep Learning framework for Intensity-based Image-to-Image Translation. 
The key idea is a new technique termed network augmentation which allows a differentiable construction of intensity histograms from images. 
We further introduce differentiable representations of (1D) cyclic and joint (2D) histograms and use them for defining loss functions based on cyclic Earth Mover's Distance (EMD) and Mutual Information (MI). While the Hue-Net can be applied to several image-to-image translation tasks, we choose to demonstrate its strength on color transfer problems, where the aim is to paint a source image with the colors of a different target image. Note that the desired output image does not exist and therefore cannot be used for supervised pixel-to-pixel learning. 
This is accomplished by using the HSV color-space and defining an intensity-based loss that is built on the EMD between the cyclic hue histograms of the output and the target images. To enforce color-free similarity between the source and the output images, we define a semantic-based loss by a differentiable approximation of the MI of these images.   
The incorporation of histogram loss functions in addition to an adversarial loss enables the construction of semantically meaningful and realistic images. 
Promising results are presented for different datasets. 
\end{abstract}
\begin{figure}[t!]
\begin{tabular}{c}
\includegraphics[width=0.32\linewidth]{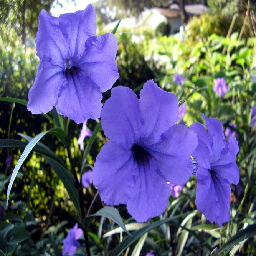} 
\includegraphics[width=0.32\linewidth]{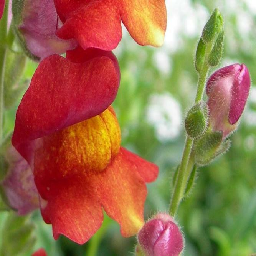} 
\includegraphics[width=0.32\linewidth]{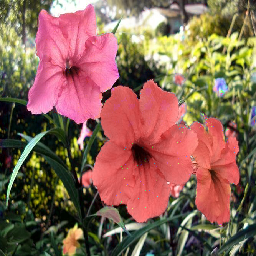} 
\\
\includegraphics[width=0.32\linewidth]{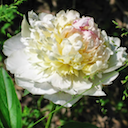} 
\includegraphics[width=0.32\linewidth]{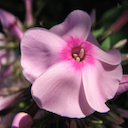} 
\includegraphics[width=0.32\linewidth]{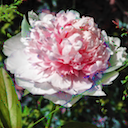} 
\\
Source\hspace{0.22\linewidth}Target\hspace{0.22\linewidth}Output
\end{tabular}
\caption{{\bf Color transfer.} The output image is obtained by painting the source image with the colors of the target image. Examples from the Oxford 102 Category Flower~\cite{Nilsback08} and the Kaggle's flower color images datasets, are shown in the first and the second rows, respectively.\label{fig:CT-main}}
\end{figure}
\vspace{-.3cm}
\section{Introduction}
\vspace{-.2cm}
Convolutional Neural Networks (CNNs) dramatically improve the state-of-the-art in many practical domains~\cite{krizhevsky2012imagenet,lecun1998gradient}. While numerous loss functions were proposed, metrics based on intensity histograms, which represent images by their gray-level distributions~\cite{Gonzalez2008,Szeliski2010} are not considered. The main obstacle seems to be histogram construction 
which is not a differentiable operation and therefore cannot be incorporated into a deep learning framework.
In this work, we introduce the Hue-Net - a deep neural network which is an image generator augmented by layers which allow, in a differentiable manner, the construction of intensity histograms of the generated images.
We further define differentiable intensity-based and semantic-based similarity measures between pairs of images that are exclusively built on their histograms.
Using these similarity measures as loss functions allows to address 'image-to-image translation'.
This paper focuses on color transfer, as shown in Figure~\ref{fig:CT-main}, where the aim is to paint a source image with the colors of a different target image. In this type of problems the desired output is rarely available and therefore, loss functions based on pixel-to-pixel comparison cannot be used. In contrast, using the proposed histogram-based loss functions the Hue-Net can be trained in an unsupervised manner, providing realistic and semantically meaningful images at test time.
\begin{figure}[t!]
\begin{centering}
\begin{tabular}{l}
\includegraphics[width=0.3\linewidth]{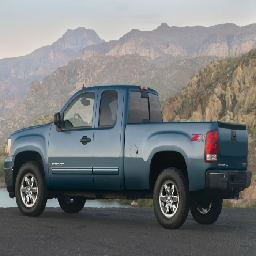}
\includegraphics[width=0.3\linewidth]{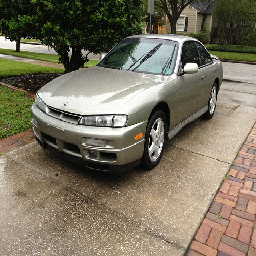}
\includegraphics[width=0.3\linewidth]{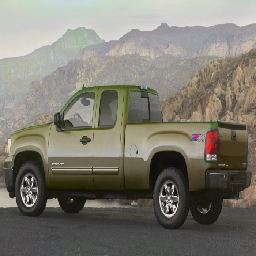}
\tabularnewline
\includegraphics[width=0.3\linewidth]{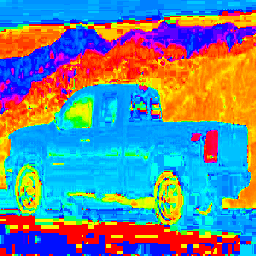}
\includegraphics[width=0.3\linewidth]{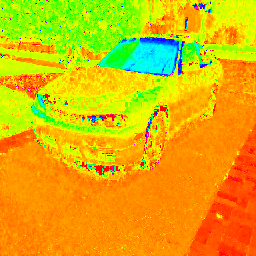}
\includegraphics[width=0.3\linewidth]{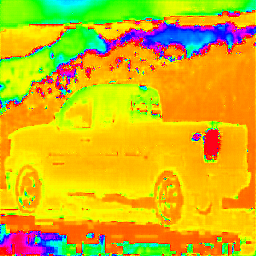}
\tabularnewline
\includegraphics[width=0.3\linewidth]{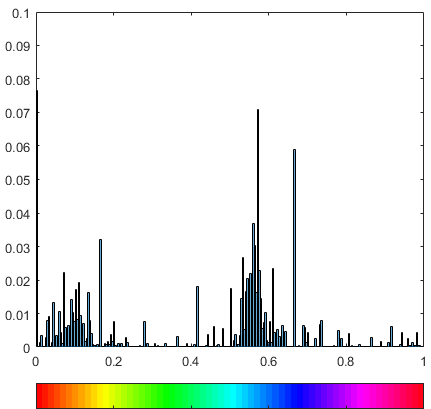}
\includegraphics[width=0.3\linewidth]{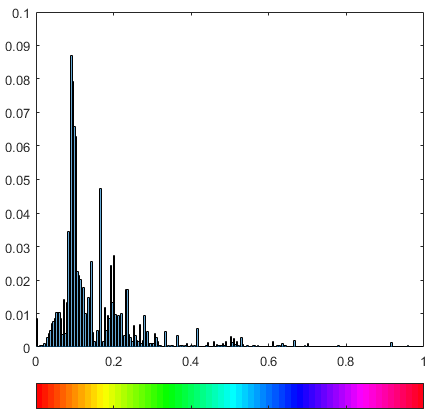}
\includegraphics[width=0.3\linewidth]{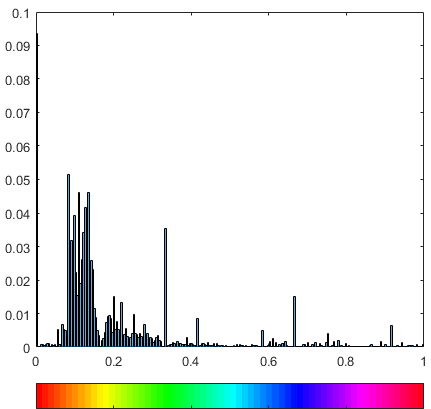}
\tabularnewline
\includegraphics[width=0.3\linewidth]{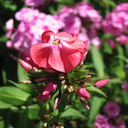}
\includegraphics[width=0.3\linewidth]{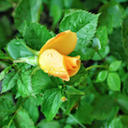}
\includegraphics[width=0.3\linewidth]{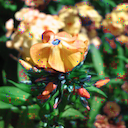}
\tabularnewline
\includegraphics[width=0.3\linewidth]{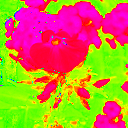}
\includegraphics[width=0.3\linewidth]{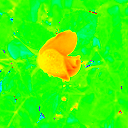}
\includegraphics[width=0.3\linewidth]{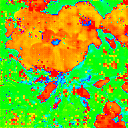}
\tabularnewline
\includegraphics[width=0.3\linewidth]{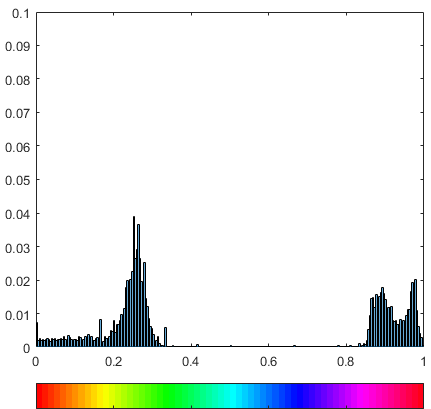}
\includegraphics[width=0.3\linewidth]{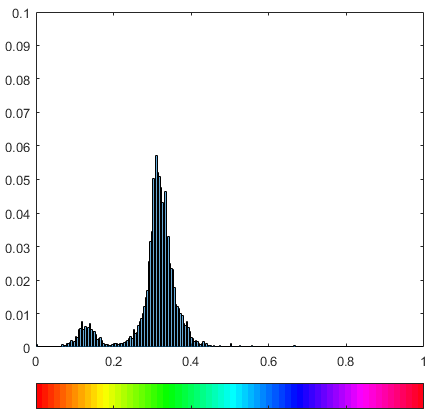}
\includegraphics[width=0.3\linewidth]{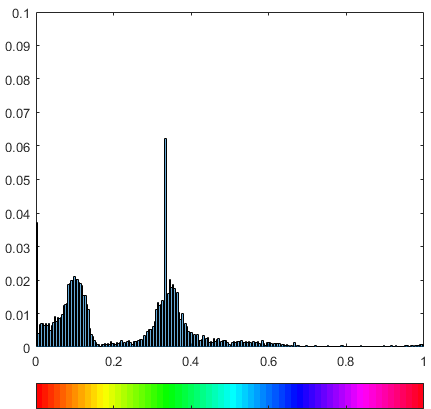}
\tabularnewline
\hspace{.8cm} Source \hspace{1.5cm} Target \hspace{1.5cm} Output \tabularnewline
\end{tabular}
\par\end{centering}
\caption{Color transfer exemplified for the flowers (rows 1-3) and cars~\cite{KrauseStarkDengFei-Fei_3DRR2013} (rows 4-6) datasets. The figure shows the source, target and output images (rows 1,4) along with the respective hue channels (rows 2,5) and hue channels' histograms (rows 3,6). The hue color-bars are presented below the histograms.}
\label{fig:CT-Hue}
\vspace{-.2cm}
\end{figure}
\begin{figure}[t!]
\includegraphics[width=0.46\linewidth]{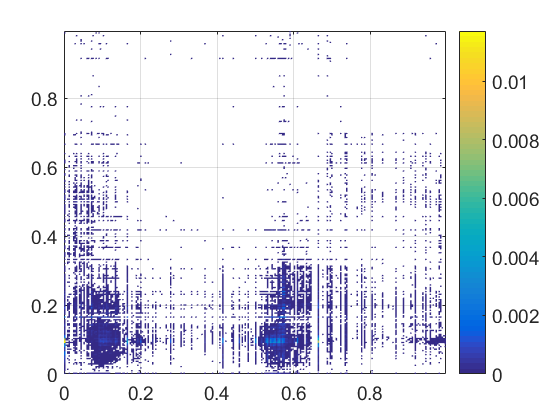}
\includegraphics[width=0.46\linewidth]{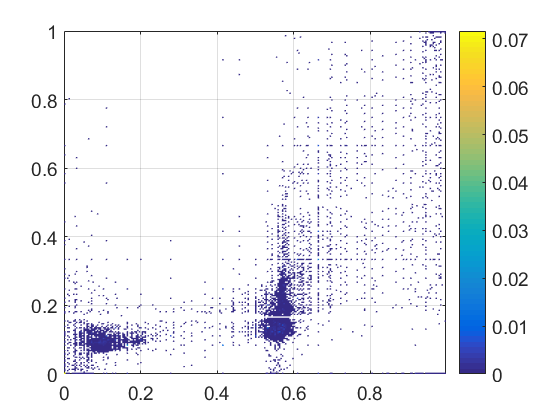}
\caption{Joint histograms ($K \times K$ matrices) of the source and target hue channels (left) and the source and the output hue channels (right) of the car images shown in Figure~\ref{fig:CT-Hue}. The values of the 2D histogram bins are color-coded. The $[0,1]$ grey-level domain is partitioned into $K=256$ bins.
\label{fig:jointHist}}
\end{figure}
An intensity histograms is a useful representation for image-to-image translation tasks. A widely used technique is histogram equalization (HE), which enhances image contrast by evening out the intensity histogram to the entire dynamic range. While its simplicity may be appealing, HE tends to provide unrealistic results as the dependency between neighboring pixels is not considered. Some works aimed to address this problem by partitioning the image space and equalizing the locally generated histograms~ \cite{Abdullah2007,Celik2011,Lee2013,PIZER1987355,Stark2000,Zuiderveld1994}.
Classical methods for color transfer were based on the concept of histogram matching, where the main idea was to adapt a color histogram of a given image to the target image. Neumann \etal ~\cite{neumann2005color} used 3D histogram matching in HSL colorspace, similarly to the sampling of multivariable functions applying a sequential chain of conditional probability density functions.
Reinhard \etal ~\cite{reinhard2001color} achieved color transfer by using a statistical analysis in Lab color space.

In this work, we exploit histogram matching using the network as an optimizer. The distance between a pair of histograms is defined by the Earth Mover's Distance (EMD). Observing that the HSV (hue, saturation, value) colorspace is most suitable for addressing color transfer problems, we use a cyclic version of the EMD when applied to the hue channel. Hue channels of some images are presented in rows 2,5 of Figure~\ref{fig:CT-Hue}.

A deformation of the intensity distribution of an image can distort its content, therefore enforcement of the semantic similarity between the source and the `color transferred' images is required. The main problem is that images have different intensities in corresponding locations making pixel-to-pixel comparison not applicable. To address this issue, we suggest to use the mutual information (MI) of the source and the output images as a measure of their color-free similarity.
In a seminal work Viola and Wells~\cite{viola1997alignment} used a cost function based on MI for image registration, where the target image and the source have different intensity distributions.
Since then, MI-based registration became popular in biomedical imaging applications, in particular when the alignment of medical images acquired by different imaging modalities is addressed.
An essential component for calculating the MI of two images is the generation of their joint histogram (see Figure~\ref{fig:jointHist}).
In the context of image registration it is called a co-occurrence matrix. While there has been significant work exploiting co-occurrence matrices, the use of joint histograms and MI for color transfer (to the best of our knowledge) has not been done before.
Moreover, differential construction of intensity histograms and joint histograms as part of a deep learning framework is done here for the first time.

Recent image generation approaches and image-to-image translation, in particular are mostly based on deep learning.
Since the main aim is generating realistic examples, adversarial frameworks, in which an adversarial network is trained on discriminating between real and fake examples, seem to be very effective~\cite{goodfellow2014generative}.
In their pix2pix framework, Isola \etal performed image-to-image translation (\eg~colorization of gray scale images) by using adversarial loss as well as $L_1$ loss between corresponding pixels in the network's output and the desired target image.
In this sense, the pix2pix is a fully supervised method and obviously cannot be applied to problems (such as color transfer) where the desired target image does not exist.
A more recent paper by Zhu \etal ~\cite{zhu2017unpaired} referred to image-to-image translation in unpaired setting using cyclic GANs.
While we agree that an adversarial loss is needed to enhance the generated images, the intensity-based loss is needed for `transferring' the source colors to that of the target. We also show (Section~\ref{sec:ExperimentalResults}) that the semantic-based loss is essential. 
Color transfer, has been recently addressed by~\cite{Mingming2017} providing visually appealing results by using neural image representation for matching. While being interesting and promising, this approach is very different then ours.

The Hue-Net is comprised of a generator, which is the well known U-Net - an encoder-decoder with skip connections~\cite{ronneberger2015u}. Yet, the main contributions are the augmented parts of the Hue-Net which allow differential construction of intensity (1D) and joint (2D) histograms, such that histogram-based loss functions are used to train the image generator in an end-to-end manner.
We demonstrate the proposed framework for color-transfer of images from three different datasets. While the results are promising we believe that the tools we developed can be applicable to different image-to-image translation tasks with slight modifications.
\vspace{-.2cm}
\section{Methods}
\vspace{-.1cm}
In this section we review the main principles we used in order to define intensity based metric via 1D histogram comparison (Section~\ref{ssec:IntensityBasedMetrics}) and semantic-based metrics via joint (2D) histogram (Section~\ref{ssec:SemanticBasedMetrics}).
The Hue-Net architecture and its augmented parts are presented in Sections~\ref{ssec:Hue-NetArchitecture}-\ref{ssec:NetworkAugmentation}, respectively. The Hue-Net loss functions are presented in Section~\ref{ssec:Hue-NetLosses}.
We conclude the section by presenting implementation details (Section~\ref{ssec:ImplementationDetails}).
\vspace{-.1cm}
\subsection{Intensity based metric}\label{ssec:IntensityBasedMetrics}
\vspace{-.1cm}
In the following section we derive a differentiable approximation to intensity histograms and present the metric we use for histogram matching.
\vspace{-.1cm}
\subsubsection{HSV Color representation}
\vspace{-.1cm}
To address the color transfer problem we choose the HSV (Hue, Saturation, Value) color space. The HSV as opposed to the RGB representation allows to separate the color component of the image (hue) from the color intensity (saturation) and the black/white ratio (value).
The proposed Hue-Net is designed to generate the hue channel ($C_{\scriptsize\mbox{H}}$) of the desired output image, while its input consists of all HSV channels of the source image and the hue channel of the target image (Figure~\ref{fig:Architecture}-d). The intensity-based metric to be hereby defined is applied to the Hue-Net output (the generated hue channel) with respect to the target's hue channel.
\vspace{-.1cm}
\subsubsection{Differentiable intensity histogram formulation}\label{sssec:differentiableintensity histogram formulation}
\vspace{-.1cm}
Images acquired by digital cameras have discrete range of $K$ intensity values. Their intensity distributions can be described with intensity histograms obtained by counting the number of pixels in each intensity value.
We, however, define a gray level image $I:\Omega\rightarrow [0,1]$ considering also generated images that can take any value in the continuous range $[0,1]$.
Let $I(x)\in[0,1]$ define a gray level value of an image pixel $x \in \Omega.$
We use the Kernel Density Estimation (KDE) for estimating the intensity density $f_I$ of an image $I$ as follows
\begin{equation}
\hat{f}_I (g) = \frac{1}{NW} \sum_{x\in\Omega} \mathcal{K}\left(\frac{I(x)-g}{W}\right)
\label{eq:IntensityDistribution}
\end{equation}
where $g\in[0,1]$, $\mathcal{K}(\cdot)$ is the kernel, $W$ is the bandwidth and $N=\lvert\Omega\rvert$ is the number of pixels in the image.
We choose the kernel $\mathcal{K}(\cdot)$ as the derivative of the logistic regression function $\sigma(z)$ as follows
\begin{equation}
\mathcal{K}(z) = \frac{d}{dz} \sigma(z) = \sigma(z)\sigma(-z)
\label{eq:kernel}
\end{equation}
where $\sigma(z)=\frac{1}{1+e^{-z}}$.
We note that Eq.~(\ref{eq:kernel}) is a non-negative real-valued integrable function and satisfies the requirements for a kernel (normalization and symmetry).

For the construction of smooth and differentiable image histogram, we partition the interval $[0,1]$ into $K$ sub intervals $\{B_k\}_{k=0}^{K-1}$, each interval with length $L=\frac{1}{K}$ and center $\mu_k = L(k+\frac{1}{2})$, then $B_k=[kL,(k+1)L]$.
We then can define the probability of pixel in the image to belong to certain intensity interval (the value of normalized histogram's bin) as
\begin{equation}
P_I(k)  \triangleq \Pr(g\in B_k) = \int_{B_k}{\hat{f}_I(g)dg}
\label{eq:IntensityProbabilityIntegral}
\end{equation}
By solving the integral we get
\begin{equation}
\begin{split}
P_I(k)
&= \frac{1}{N} \sum_{x\in\Omega} \sigma\left(\frac{I(x)-g}{W}\right)\Big|^{\mu_k - L/2}_{\mu_k+L/2} \\
&= \frac{1}{N} \sum_{x\in\Omega} \Bigl[\sigma\left(\frac{I(x)-\mu_k+L/2}{W}\right)\\
&\qquad\qquad -\sigma\left(\frac{I(x)-\mu_k-L/2}{W}\right)\Bigr]
\end{split}
\label{eq:IntensityProbabilityRect}
\end{equation}
The function $P_I(k)$ which provides the value of the $k^{\mbox{\small{th}}}$ bin in a differentiable histogram
can be rewritten as follows:
\begin{equation}
P_I(k) = \frac{1}{N}\sum_{x\in\Omega} \Pi_k(I(x)),
\label{eq:IntensityProbabilityFinal}
\end{equation}
where,
\begin{equation}
\Pi_k(z) \triangleq \sigma(\frac{z-\mu_k + L/2}{W})-\sigma(\frac{z-\mu_k - L/2}{W})
\label{eq:Pi_k}
\end{equation}
is a differentiable approximation of the Rect function.
Finally, we define the differentiable histogram $\mathbf{h}$ of an image $I$ as follows
\begin{equation}
\mathbf{h}=\{\mu_k,P_{I}(k)\}_{k=0}^{K-1}
\label{eq:Histogram}
\end{equation}
\vspace{-.1cm}
\subsubsection{Earth Mover's Distance}\label{sssec:EarthMoversDistance}
\vspace{-.1cm}
We use the EMD~\cite{rubner2000earth}, also known as the Wasserstein metric~\cite{dobrushin1970prescribing} to define the distance between two histograms.
Let $\mathbf{h_1}$ and $\mathbf{h_2}$ be the source and the target histograms of images $I_1$ and $I_2$ , respectively:
\begin{equation}
\mathbf{h}_j=\{\mu_{k_j},P_{I_j}(k_j)\}_{k_j=0}^{K-1}
\end{equation} where $j=1,2$.
Let $\mathbf{D}$ be the ground distance matrix, where its $k_1, k_2$-th entry $d_{k_1,k_2}$ is the ground distance between $\mu_{k_1}$ and $\mu_{k_2}.$ 
Let $\mathbf{F}$ be the transportation matrix, where its $k_1, k_2$-th entry $f_{k_1,k_2}$ indicates the mass transported from $\mu_{k_1}$ to $\mu_{k_2}$.
We aim to find a flow $\mathbf{F}$ that minimizes the overall cost:
\begin{equation}
W(\mathbf{h_1}, \mathbf{\mathbf{h_2}}, \mathbf{F})= \sum_{k_1=0}^{K-1}\sum_{k_2=0}^{K-1} d_{k_1,k_2} f_{k_1,k_2}
\label{eq:EMD-Work}
\end{equation}
subject to the following constraints:
\begin{equation}
\begin{aligned}
&f_{k_1,k_2} \geq 0 \quad \textrm{for all } k_1,k_2 \\
&\sum_{k_2=0}^{K-1} f_{k_1,k_2} \leq P_{I_1}(k_1) \quad \textrm{for all } k_1 \\
&\sum_{k_1=0}^{K-1} f_{k_1,k_2} \leq P_{I_2}(k_2) \quad \textrm{for all } k_2 \\
&\sum_{k_1=0}^{K-1} \sum_{k_2=0}^{K-1} f_{k_1,k_2} \leq 
\min
\Bigl(
\sum_{k_1=0}^{K-1}P_{I_1}(k_1),\sum_{k_2=0}^{K-1}P_{I_2}(k_2)
\Bigr)
\end{aligned}
\label{eq:constraints}
\end{equation}
The EMD between the two histograms is the minimum cost of work (Eq.~\ref{eq:EMD-Work}) that satisfies the constraints (Eq.~\ref{eq:constraints}) normalized by the total flow:
\begin{equation}
\mathcal{D}_{\scriptsize{\mbox{EMD}}}(\mathbf{h_1}, \mathbf{h_2}) = \inf_{\mathbf{F}} \frac{W(\mathbf{h_1}, \mathbf{h_2}, \mathbf{F})}
{\sum_{k_1=0}^{K-1}\sum_{k_2=0}^{K-1}f_{k_1,k_2}}
\label{eq:EMD-inf}
\end{equation}
For one-dimensional histograms with equal areas,
EMD has been shown to be equivalent to Mallows distance
which has the following closed-form solution~\cite{ELevina2001}:
\begin{equation}
\mathcal{D}_{\scriptsize{\mbox{EMD}}}(\mathbf{h_1}, \mathbf{h_2}) = \left(\frac{1}{K}\right)^{\frac{1}{l}} \lVert \mbox{CDF}(\mathbf{h_1}) - \mbox{CDF}(\mathbf{h_2}) \rVert _l,
\label{eq:EMD-cdf}
\end{equation}
where, $\mbox{CDF}(\cdot)$ is the cumulative density function.
We use $l = 2$ for Euclidean distance. Dropping the normalization term, we obtain an intensity-based metric:
\begin{equation}
\mathcal{D}_{\scriptsize{\mbox{EMD}^2}}(\mathbf{h_1}, \mathbf{h_2}) = \sum_{i=0}^{K-1}{\left(\mbox{CDF}_i(\mathbf{h_1})-\mbox{CDF}_i(\mathbf{h_2})\right)^2},
\label{eq:EMD2-cdf}
\end{equation}
where, $\mbox{CDF}_i(\mathbf{h_1})$ is the $i$-th element of the CDF of $\mathbf{h_1}$.
Following Hou \etal~\cite{LeHou2016} we use $\mathcal{D}_{\scriptsize{\mbox{EMD}^2}}$ instead of $\mathcal{D}_{\scriptsize{\mbox{EMD}}}$ because it usually converges faster and is easier to optimize with gradient descent~\cite{luenberger1984linear,shalev2011stochastic}.
\vspace{-.1cm}
\subsubsection{Cyclic histogram}\label{sssec:CyclicHistogram}
\vspace{-.1cm}
The hue channel in HSV colorspace is cyclic, thus the EMD measure should be adapted accordingly.
Werman \etal ~\cite{WERMAN1985} showed that the EMD is equal to the $L_1$ distance between the cumulative histograms. They also proved that matching two cyclic histograms by only examining cyclic permutations is optimal.
Therefore, the cyclic $\mathcal{D}_{\scriptsize{\mbox{EMD}^2_c}}$ distance can be expressed as
\begin{equation}
\mathcal{D}_{\scriptsize{\mbox{EMD}^2_c}} (\mathbf{h_1}, \mathbf{h_2}) =  \min_{l=0,\dots,K-1} \mathcal{D}_{\scriptsize{\mbox{EMD}^2_c}}(T(\mathbf{h_1}, l), T(\mathbf{h_2}, l))
\label{eq:EMD_c}
\end{equation}
where $T(\mathbf{h_j},l)$ shifted negatively the elements in $h_j$ by the offset of $l$. Elements that passed the last position will wrap around to the first. This operator can be described as
\begin{equation}
T(\mathbf{h_j},l)[k] = \mathbf{h_j}[(k+l)\bmod{K} ]
\label{eq:CyclicTransformation}
\end{equation}
\vspace{-.1cm}
\subsection{Semantic-based metric}\label{ssec:SemanticBasedMetrics}
\vspace{-.1cm}
In a color transfer problem, we wish to constrain semantic similarity between the source image $I_1$ and the color transformed output image $I_2$, where each has a different intensity distribution. This is accomplished by applying the semantic-based metric, to be hereby defined, to the generated hue channel with respect to the source's hue channel (Figure~\ref{fig:Architecture}-k).
This metric will be based on the MI between these hue channels.
\vspace{-.1cm}
\subsubsection{Mutual information}\label{sssec:Mutual information}
\vspace{-.1cm}
The MI of two images $I_1$ and $I_2$ is defined as follows:
\begin{equation}
\begin{aligned}
\mathcal{I}(I_1, I_2) = \sum_{k_1=0}^{K-1} \sum_{k_2=0}^{K-1} P_{I_1,I_2}(k_1,k_2)\log{\frac{P_{I_1,I_2}(k_1, k_2)}{P_{I_1}(k_1)P_{I_2}(k2)}},
\end{aligned}
\label{eq:MI}
\end{equation}
where, $P_{I_1}$ ,$P_{I_2}$ are the image histograms as defined is Eq.~\ref{eq:IntensityProbabilityFinal}, and $P_{I_1,I_2}$ is the joint histogram that will be described next section.
Maximizing the MI between the output and the input image allows us to generate images which are semantically similar. Following~\cite{alex2003hierarchical} we define the semantic-based loss as follows:
\begin{equation}
\mathcal{D}_{\scriptsize\mbox{MI}}
(I_1,I_2) = 1 - \frac{\mathcal{I}(I_1,I_2)}
{\mathcal{H}(I_1,I_2)},
\label{eq:D_MI}
\end{equation}
where, $\mathcal{H}(I_1,I_2)$ is the joint entropy of $I_1$, $I_2$ defined as
\begin{equation}
\mathcal{H}(I_1,I_2)=- \sum_{k_1=0}^{K-1} \sum_{k_2=0}^{K-1} P_{I_1,I_2}(k_1,k_2)\log{P_{I_1,I_2}(k_1,k_2)}.
\label{eq:entropy}
\end{equation}
The quantity $\mathcal{D}(I_1,I_2)$ is a metric~\cite{alex2003hierarchical}, with $\mathcal{D}(I_1, I_1) = 0$ and $\mathcal{D}(I_1, I_2) \leq 1$ for all pairs $(I_1,I_2)$. This metric has symmetry, positivity and boundedness properties.
\vspace{-.1cm}
\subsubsection{Differentiable joint intensity histogram}\label{sssec:JointIntensityHistogramFormulation}
\vspace{-.1cm}
We use the multivariate KDE for estimating the joint intensity density $f_{I_1,I_2}$ of two images $I_1,I_2:\Omega\rightarrow [0,1]$, as follows:
\begin{equation}
\hat{f}_{I_1,I_2} (g_1,g_2) = \frac{1}{N}
\lvert\mathbf{W}\rvert^{-1/2}
\sum_{x\in\Omega}
\mathcal{K} \left( \mathbf{W}^{-1/2} (\mathbf{I}(x) - \mathbf{g}) \right)
\label{eq:JointIntensityDistribution}
\end{equation}
where, $\mathbf{I}(x) = \begin{bmatrix} I_1 & I_2
\end{bmatrix}^T$, $\mathbf{g} = \begin{bmatrix} g_1 & g_2
\end{bmatrix}^T$, $\mathbf{W}$ is the bandwidth (or smoothing) $2\times 2$ matrix and $\mathcal{K}(\cdot,\cdot)$ is the symmetric 2D kernel function.
As in the 1D case (Eq.~\ref{eq:kernel}), we choose the kernel $\mathcal{K}(\cdot,\cdot)$ as the derivative of the logistic regression function $\sigma(z)$ for each of the two variables separately:
\begin{equation}
\mathcal{K}(z_1,z_2) = \frac{d}{dz}\sigma(z_1)  \frac{d}{dz}\sigma(z_2) = \sigma(z_1)\sigma(-z_1)\sigma(z_2)\sigma(-z_2)
\label{Kernel2D}
\end{equation}
We define the bandwidth matrix $\mathbf{W}$ as
$\begin{bmatrix}
W & 0 \\ 0 & W
\end{bmatrix}$.
We define the probability of corresponding pixels in $I_1$ and $I_2$ to belong to the intensity intervals $B_{k_1}$ and $B_{k_2},$ correspondingly, as follows:
\begin{equation}
\begin{split}
P_{I_1,I_2}(k_1,k_2) \triangleq
\Pr(I_1(x)\in B_{k_1}, I_2(x)\in  B_{k_2}) \\
= \int_{B_{k_1}}\int_{B_{k_2}}{\hat{f}_{I_1,I_2}(g_1,g_2)dg_1dg_2}
\end{split}
\label{eq:JointProabilityIntegreal}
\end{equation}
By solving the integral we get:
\begin{equation}
\begin{split}
P_{I_1,I_2}(k_1,k_2) =
\frac{1}{N}\sum_{x\in\Omega}
\sigma\left(\frac{I_1(x)-g_1}{W}\right)
\Big|^{\mu_{k_1} - L/2}_{\mu_{k_1}+L/2} \\
\times\sigma\left(\frac{I_2(x)-g_2}{W}\right)
\Big|^{\mu_{k_2} - L/2}_{\mu_{k_2}+L/2}
\end{split}
\label{JointProabilityRect}
\end{equation}


By using the definition of $\Pi_k$ from Eq.~\ref{eq:Pi_k}, we can expressed the value of joint histogram $k_1,k_2$-th bin as
\begin{equation}
{P}_{I1,I2}(k_1,k_2) = \frac{1}{{N}}\sum_{x\in\Omega}{\Pi}_{k_1}(I_1(x)){\Pi}_{k_2}(I_2(x))
\label{eq:JointProabilityFinal}
\end{equation}
This equation can be also written using matrix notation.
We define a $K\times N$ matrix $\mathbf{P_j}$ where $\mathbf{P_j}_{k_j,n}\triangleq {\Pi}_{k_j}(I_j(n))$, where $n$ is the pixel index in a flatten image $I_j$, $j=1,2$.
The approximated joint histogram $\mathbf{J}$ ($K\times K $ matrix) of two images $I_1$, $I_2$ can be defined with a matrix multiplication:
\begin{equation}
\mathbf{J}(I_1,I_2) = \frac{1}{N} \mathbf{P_1} \mathbf{P_2}^T
\label{eq:JointProabilityMatrix}
\end{equation}
\vspace{-.2cm}
\subsection{Hue-Net architecture}\label{ssec:Hue-NetArchitecture}
\vspace{-.1cm}
\begin{figure*}[t!]
\begin{center}
\includegraphics[width=\linewidth]{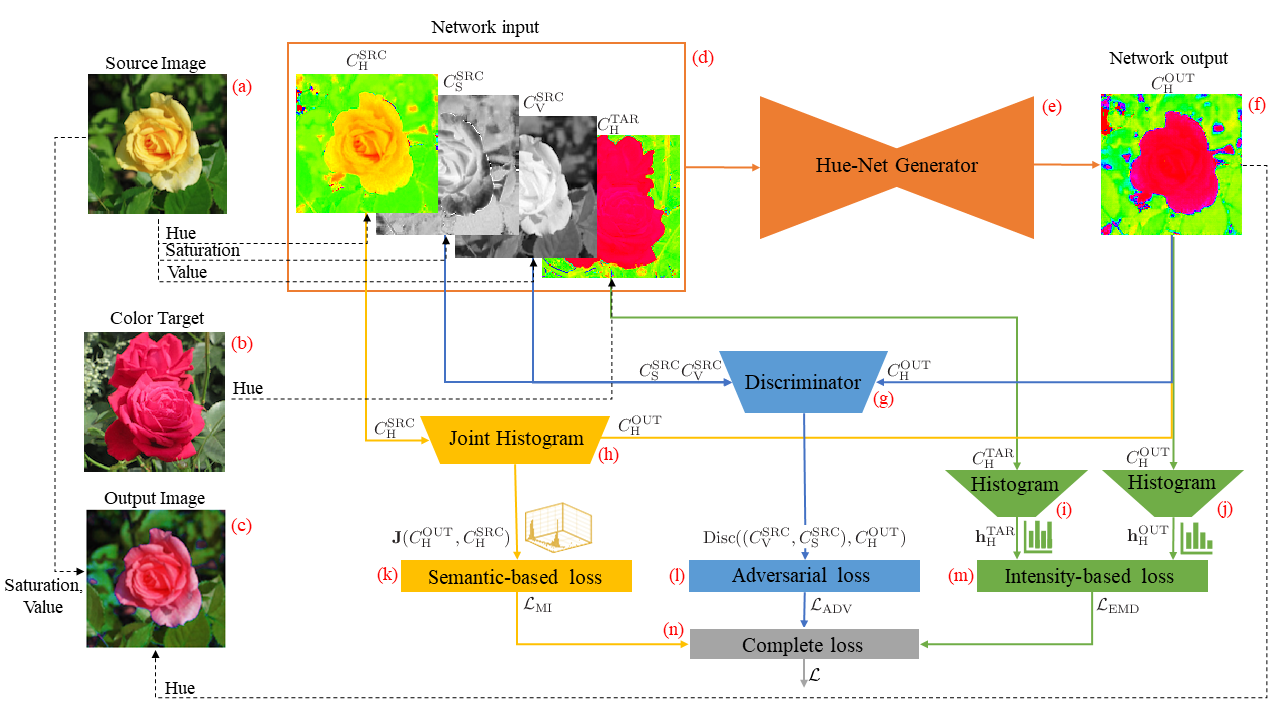}
\end{center}\caption{\textbf{Hue-Net architecture}. For the color transfer problem, we have source (a) and target (b) images. The Hue-Net input is composed of the three HSV channels of the input image as well as the hue channel of the target image (d). The Hue-Net Generator (e) generates a hue channel (f) which ``transfer'' the source hue channel. The Generator is augmented and its output is fed into three additional networks for histogram construction and intensity-based loss (green); for joint histogram construction and semantic-based loss (yellow) and for the generation of an adversarial loss (blue). The histogram construction network(j) is illustrated in detail in Figure~4. Its output, \ie~the intensity histogram of the generated hue-channel is compared to the histogram of the target hue channel (that is generated in the same manner, j) using EMD, thus defining the intensity-based losses (m).
Both the source and the generated hue channels are the input for the joint histogram construction network (h).
Its output is used for the calculation of the semantic-based loss which is a function of the mutual information between these hue-channels (k). The input to the conditional discriminator (g) consists of the generated hue channel, the source saturation and value channels (which together build the output `color-transferred' image). Its output defines the Hue-Net adversarial loss (l). The complete Hue-Net loss (n) is a weighted sum of the intensity-based, semantic-based and the adversarial losses. }
\label{fig:Architecture}
\end{figure*}
Figure~\ref{fig:Architecture} illustrates the Hue-Net architecture.
As in~\cite{IsolaZZE16}, we use the U-Net architecture~\cite{ronneberger2015u} for the Hue-Net generator and the convolutional “PatchGAN” classifier~\cite{li2016precomputed} for its discriminator. 
The U-Net generates the hue channel of the output image.
The augmented parts of the Hue-Net are detailed next.
\vspace{-.1cm}
\subsection{Network augmentation}\label{ssec:NetworkAugmentation}
\vspace{-.1cm}
\subsubsection{1D Histogram construction}\label{sssec:1D Histogram construction}
\vspace{-.1cm}
\begin{figure}[h]
\begin{center}
\includegraphics[width=7cm]{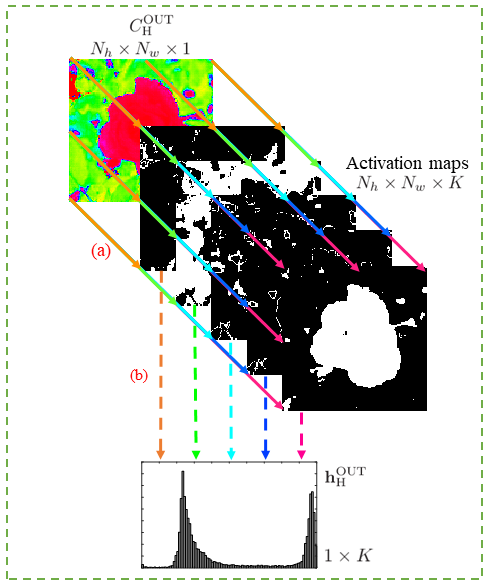}
\end{center}
\caption{\textbf{Histogram construction} network augmentation for histogram construction, we apply $\{\Pi_k(\cdot)\}_{k=0}^{k=K-1}$ for all pixels in output image (a)
$C^{\scriptsize{\mbox{OUT}}}_{\scriptsize{\mbox{H}}}$,
and get $K$ channels (Activation maps), the elements of each channel are summed (b) for the construction of the histogram.}
\label{fig:HistogramConstruction}
\end{figure}
For the construction of an intensity histogram the output hue channel (of size $N_h\times N_w$ )
 is replicated $K$ times, where each pixel is represented by a smooth 1-hot vector.
The $k$-th channel is obtained by an application of the function $\Pi_k(z)$ to the output's pixels, thus representing its `contribution' to the $k$-th histogram bin (in the interval $B_k$).
An intensity histogram is constructed by a summation of each of these $K$ channels (Eq.~\ref{eq:IntensityProbabilityFinal}).
This operation is illustrated in Figure~\ref{fig:HistogramConstruction}.

We construct cyclic permutations of the constructed histogram by matrix multiplication of $\mathbf{h}_j$ and circulant matrix (a special kind of Toeplitz matrix), to obtain the transformation described in Eq.~\ref{eq:CyclicTransformation}.
\vspace{-.1cm}
\subsubsection{Joint Histogram construction}\label{sssec:JointHistogramConstruction}
\vspace{-.1cm}
Similar to the construction of the 1D histograms, the generation of the joint histogram of two images $I_1$ and $I_2$ is based on the construction of $K$ channels from each. Each channel is reshaped into an $1\times N$ vector (activation map in Figure~\ref{fig:HistogramConstruction}) , where $N =N_h\times N_w$ is the image size in pixels. The K-channels form a $K\times N$ matrix, which is exactly $\mathbf{P_j}$ as defined in Eq.~\ref{eq:JointProabilityMatrix}. Multiplying and normalizing the matrices obtained for each image we get the joint histogram $J(I_1,I_2)$ as defined in Eq.~\ref{eq:JointProabilityMatrix}.

\vspace{-.1cm}
\subsection{Hue-Net losses}\label{ssec:Hue-NetLosses}
\vspace{-.1cm}
The complete Hue-Net loss $\pazocal{L}$ is a weighted sum of three loss functions:
\begin{equation}
\pazocal{L} =
\lambda\pazocal{L}_{\scriptsize\mbox{EMD}}
+ \gamma\pazocal{L}_{\scriptsize\mbox{MI}} +
\pazocal{L}_{\scriptsize\mbox{ADV}}
\label{eq:LossAll}
\end{equation}
where $\pazocal{L}_{\scriptsize\mbox{EMD}}$, $\pazocal{L}_{\scriptsize\mbox{MI}}$, $\pazocal{L}_{\scriptsize\mbox{ADV}}$
are the intensity-based loss using EMD, the semantic-based loss using MI and the adversarial loss, respectively. The scalars $\lambda$, $\gamma$ are the weights.
Note, that we empirically set $\lambda$=100, $\gamma$=25.\\
{\bf Intensity-based loss} \\
The intensity-based loss is derived from Eq.~\ref{eq:EMD_c} which defines the EMD between the two cyclic histograms ${\mathbf{h}}_{\scriptsize\mbox{H}}^{\scriptsize\mbox{TAR}},
{\mathbf{h}}_{\scriptsize\mbox{H}}^{\scriptsize\mbox{OUT}}$ of the target and the output:
\begin{equation}
\pazocal{L}_{\scriptsize\mbox{EMD}} = \mathcal{D}_{\scriptsize{\mbox{EMD}^2_c}}
(
{\mathbf{h}}_{\scriptsize\mbox{H}}^{\scriptsize\mbox{TAR}},
{\mathbf{h}}_{\scriptsize\mbox{H}}^{\scriptsize\mbox{OUT}}
)
\label{eq:LossIntensity}
\end{equation}
{\bf Semantic-based loss}
\\
Semantic-based loss between the hue channels of the network's output $C^{\scriptsize{\mbox{OUT}}}_{\scriptsize{\mbox{H}}}$ and the source image $ C^{\scriptsize{\mbox{SRC}}}_{\scriptsize{\mbox{H}}}$
is based on their MI (Eq.~\ref{eq:D_MI}) and defined as follows:
\begin{equation}
\pazocal{L}_{\scriptsize{\mbox{{MI}}}} = \mathcal{D}_{\scriptsize{\mbox{MI}}}(C^{\scriptsize{\mbox{OUT}}}_{\scriptsize{\mbox{H}}},C^{\scriptsize{\mbox{SRC}}}_{\scriptsize{\mbox{H}}})
\label{eq:LossSemantic}
\end{equation}
{\bf Adversarial loss}
\\
We use conditional GAN loss similar~\cite{IsolaZZE16}.
The discriminator learns to distinguish between $C^{\scriptsize{\mbox{SRC}}}_{\scriptsize{\mbox{H}}}$  and $C^{\scriptsize{\mbox{OUT}}}_{\scriptsize{\mbox{H}}}$ conditioned by $(C^{\scriptsize{\mbox{SRC}}}_{\scriptsize{\mbox{S}}},
C^{\scriptsize{\mbox{SRC}}}_{\scriptsize{\mbox{V}}}).$
The adversarial loss for the generator is based on the discriminator output:
\begin{equation}
\pazocal{L}_{\scriptsize\mbox{ADV}} =
 -\log\left[{\mbox{Disc}((
 C^{\scriptsize{\mbox{SRC}}}_{\scriptsize{\mbox{V}}},
 C^{\scriptsize{\mbox{SRC}}}_{\scriptsize{\mbox{S}}}
 ) ,
 C^{\scriptsize{\mbox{OUT}}}_{\scriptsize{\mbox{H}}}
 )}\right]
\label{LossAdversarial}
\end{equation}
The discriminator loss is defined as:
\begin{equation}
\begin{aligned}
\pazocal{L}_{\scriptsize\mbox{DISC}} & = -[\log\left[{\mbox{Disc}((
 C^{\scriptsize{\mbox{SRC}}}_{\scriptsize{\mbox{V}}},
 C^{\scriptsize{\mbox{SRC}}}_{\scriptsize{\mbox{S}}}
 ) ,C^{\scriptsize{\mbox{SRC}}}_{\scriptsize{\mbox{H}}} ))}\right]+ \\
 &    \log{[
 1-\mbox{Disc}((C^{\scriptsize{\mbox{SRC}}}_{\scriptsize{\mbox{V}}},
 C^{\scriptsize{\mbox{SRC}}}_{\scriptsize{\mbox{S}}}) ,
 C^{\scriptsize{\mbox{OUT}}}_{\scriptsize{\mbox{H}}} ))]]}
\end{aligned}
\label{LossDiscriminator}
\end{equation}
where $\mbox{Disc}(\cdot,\cdot)$ is the discriminator output.
\vspace{-.2cm}
\subsection{Implementation Details}\label{ssec:ImplementationDetails}
\vspace{-.1cm}
To optimize our networks, we alternate between two gradient descent steps training the Generator (G) and the Discriminator (D).
As suggested in~\cite{goodfellow2014generative}, we train $G$ to maximize $log D(x, G(x, z)).$
We use a batch size of $32$, minibatch SGD and apply the Adam solver~\cite{kingma2014adam}, with a learning rate of $0.0002$, and momentum parameters $\beta_1 = 0.5$, $\beta_2 = 0.999$.
We augmented the training images with vertical flips.
We randomly shuffled the dataset to get different pairs of source and target images in each epoch. Histograms were constructed using $K=256$ bins.
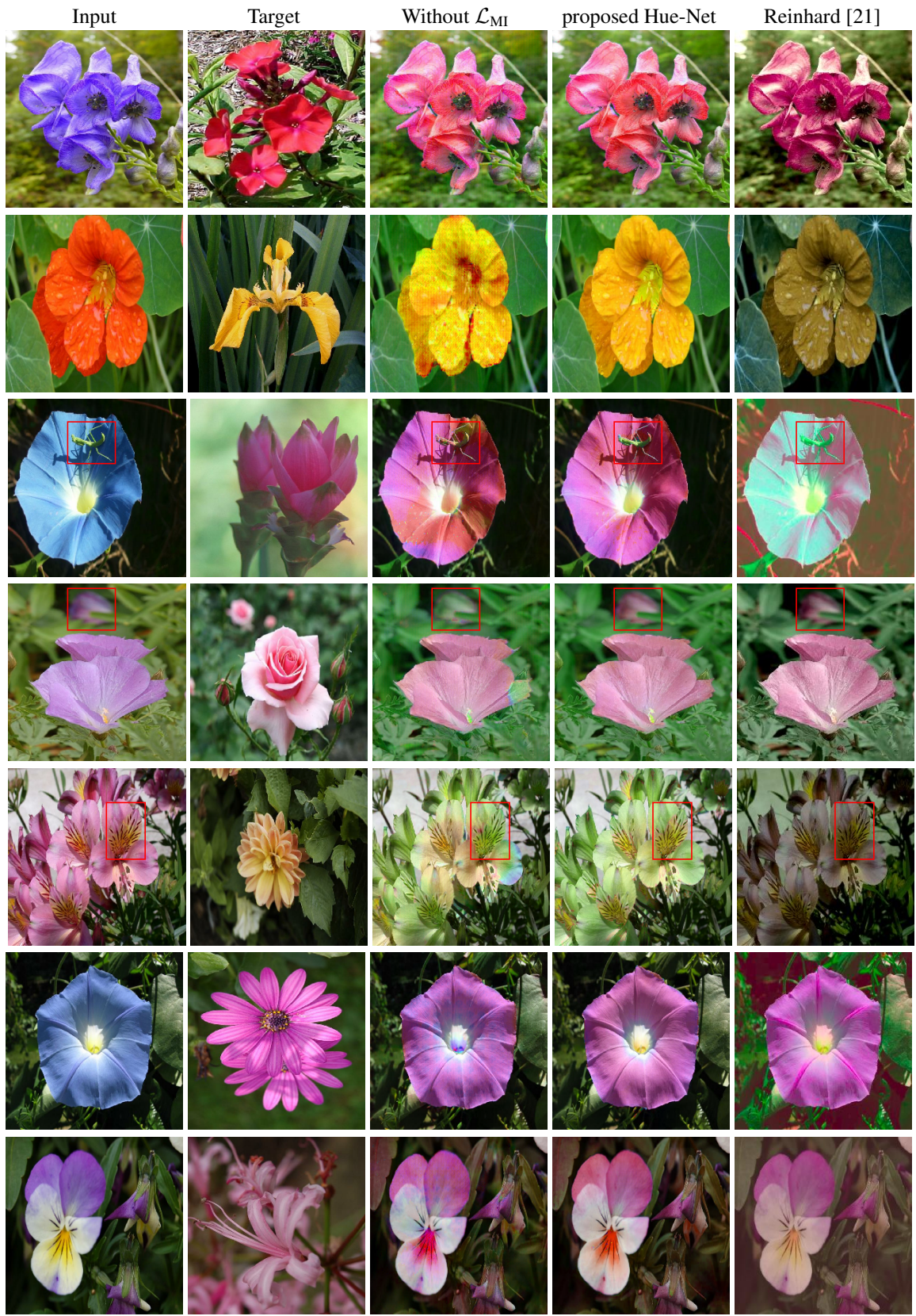
\begin{figure*}
\begin{tabular}{c}
\hspace{1.2cm}\hspace{0.6cm}Input\hspace{2.2cm}Target\hspace{1.7cm}Without $\pazocal{L}_{\scriptsize\mbox{MI}}$  \hspace{0.8cm}proposed Hue-Net\hspace{0.8cm}Reinhard~\cite{reinhard2001color}\\
\hspace{1.2cm}\includegraphics[width=0.17\linewidth]{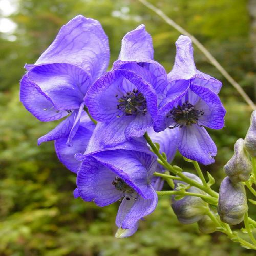}
\includegraphics[width=0.17\linewidth]{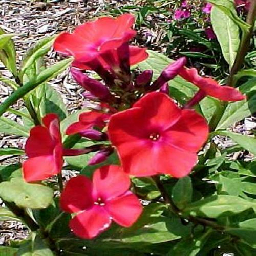}
\includegraphics[width=0.17\linewidth]{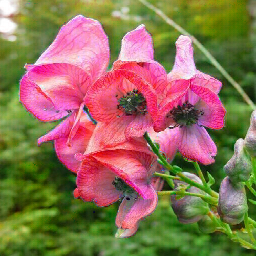}
\includegraphics[width=0.17\linewidth]{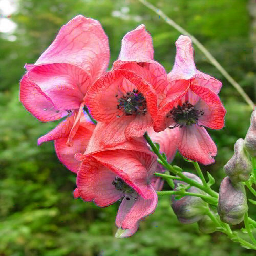}
\includegraphics[width=0.17\linewidth]{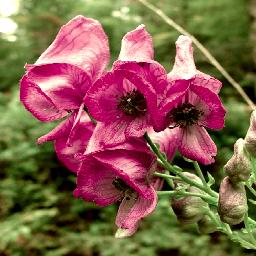} \\
\hspace{1.2cm}\includegraphics[width=0.17\linewidth]{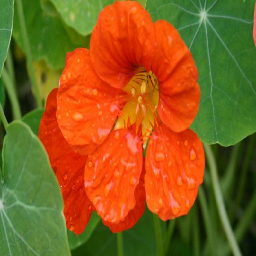}
\includegraphics[width=0.17\linewidth]{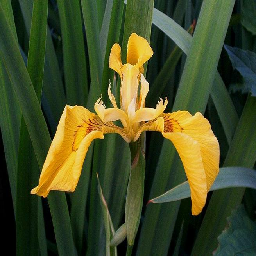}
\includegraphics[width=0.17\linewidth]{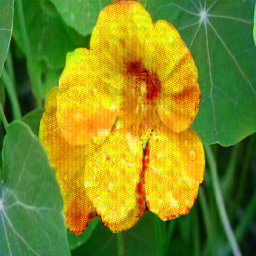}
\includegraphics[width=0.17\linewidth]{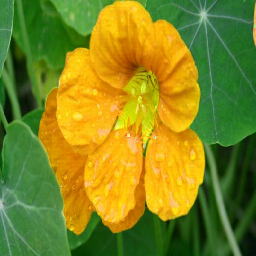}
\includegraphics[width=0.17\linewidth]{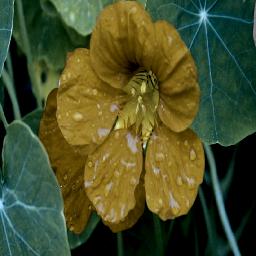} \\
\hspace{1.2cm}
\begin{tikzpicture}
    \node[anchor=south west,inner sep=0] at (0,0) {\includegraphics[width=0.17\linewidth]{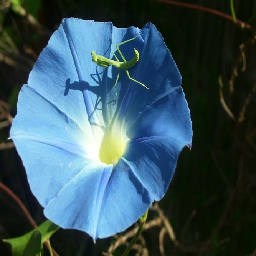}};
    \draw[red,semithick] (1,1.9) rectangle (1.8,2.6);
\end{tikzpicture}
\includegraphics[width=0.17\linewidth]{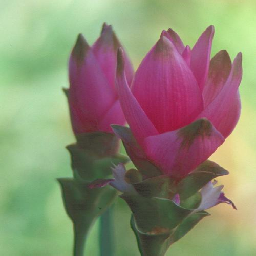}
\begin{tikzpicture}
    \node[anchor=south west,inner sep=0] at (0,0) {\includegraphics[width=0.17\linewidth]{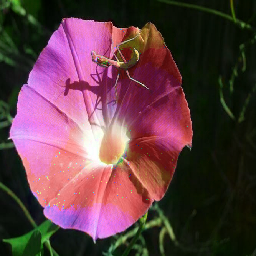}};
    \draw[red,semithick] (1,1.9) rectangle (1.8,2.6);
\end{tikzpicture}
\begin{tikzpicture}
    \node[anchor=south west,inner sep=0] at (0,0) {\includegraphics[width=0.17\linewidth]{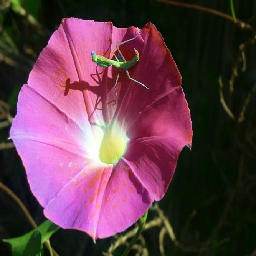}};
    \draw[red,semithick] (1,1.9) rectangle (1.8,2.6);
\end{tikzpicture}
\begin{tikzpicture}
    \node[anchor=south west,inner sep=0] at (0,0) {\includegraphics[width=0.17\linewidth]{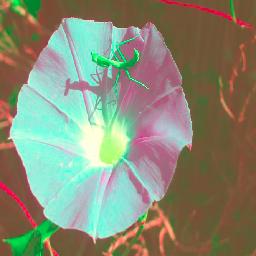}};
    \draw[red,semithick] (1,1.9) rectangle (1.8,2.6);
\end{tikzpicture}
\\
\hspace{1.2cm}
\begin{tikzpicture}
    \node[anchor=south west,inner sep=0] at (0,0) {\includegraphics[width=0.17\linewidth]{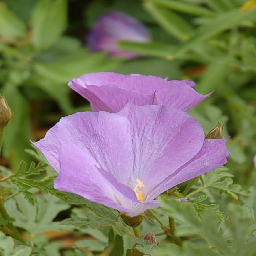}};
    \draw[red,semithick] (1,2.2) rectangle (1.8,2.9);
\end{tikzpicture}
\includegraphics[width=0.17\linewidth]{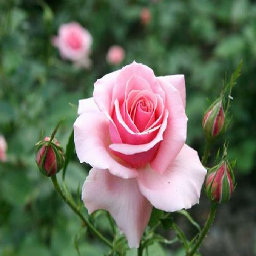}
\begin{tikzpicture}
    \node[anchor=south west,inner sep=0] at (0,0) {\includegraphics[width=0.17\linewidth]{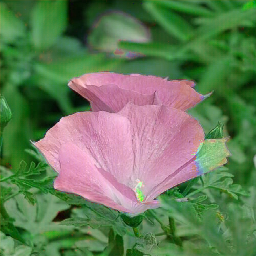}};
    \draw[red,semithick] (1,2.2) rectangle (1.8,2.9);
\end{tikzpicture}
\begin{tikzpicture}
    \node[anchor=south west,inner sep=0] at (0,0) {\includegraphics[width=0.17\linewidth]{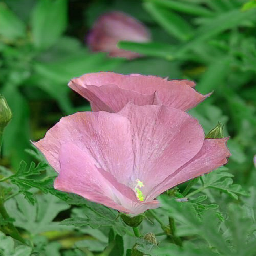}};
    \draw[red,semithick] (1,2.2) rectangle (1.8,2.9);
\end{tikzpicture}
\begin{tikzpicture}
    \node[anchor=south west,inner sep=0] at (0,0) {\includegraphics[width=0.17\linewidth]{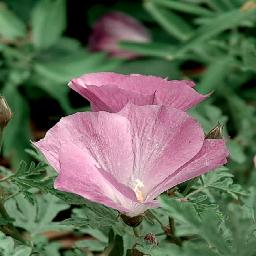}};
    \draw[red,semithick] (1,2.2) rectangle (1.8,2.9);
\end{tikzpicture}
\\
\hspace{1.2cm}
\begin{tikzpicture}
    \node[anchor=south west,inner sep=0] at (0,0) {\includegraphics[width=0.17\linewidth]{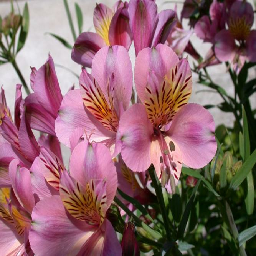}};
    \draw[red,semithick] (1.65,1.45) rectangle (2.3,2.4);
\end{tikzpicture}
\includegraphics[width=0.17\linewidth]{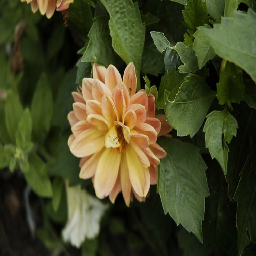}
\begin{tikzpicture}
    \node[anchor=south west,inner sep=0] at (0,0) { \includegraphics[width=0.17\linewidth]{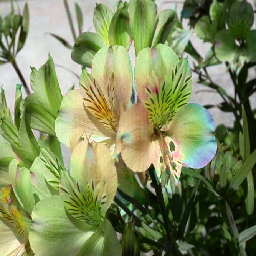}};
    \draw[red,semithick] (1.65,1.45) rectangle (2.3,2.4);
\end{tikzpicture}
\begin{tikzpicture}
    \node[anchor=south west,inner sep=0] at (0,0) { \includegraphics[width=0.17\linewidth]{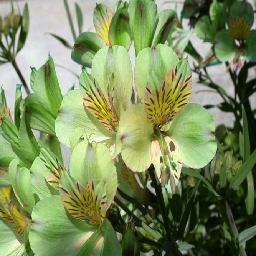}};
    \draw[red,semithick] (1.65,1.45) rectangle (2.3,2.4);
\end{tikzpicture}
\begin{tikzpicture}
    \node[anchor=south west,inner sep=0] at (0,0) {  \includegraphics[width=0.17\linewidth]{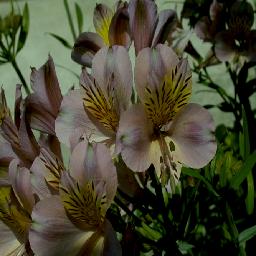}};
    \draw[red,semithick] (1.65,1.45) rectangle (2.3,2.4);
\end{tikzpicture}
 \\
\hspace{1.2cm}\includegraphics[width=0.17\linewidth]{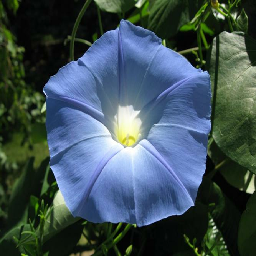}
 \includegraphics[width=0.17\linewidth]{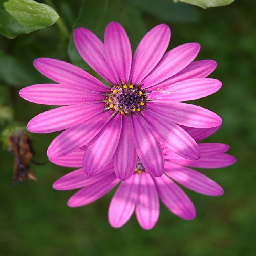}
 \includegraphics[width=0.17\linewidth]{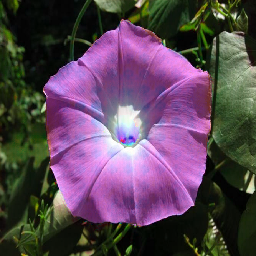}
 \includegraphics[width=0.17\linewidth]{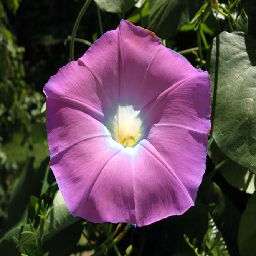}
 \includegraphics[width=0.17\linewidth]{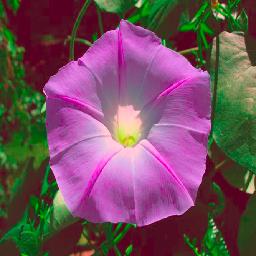} \\
\hspace{1.2cm}\includegraphics[width=0.17\linewidth]{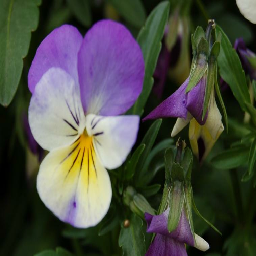}
 \includegraphics[width=0.17\linewidth]{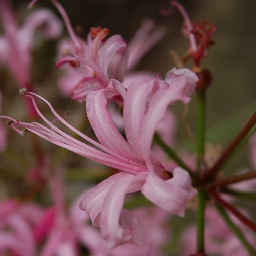}
 \includegraphics[width=0.17\linewidth]{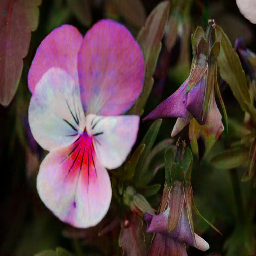}
 \includegraphics[width=0.17\linewidth]{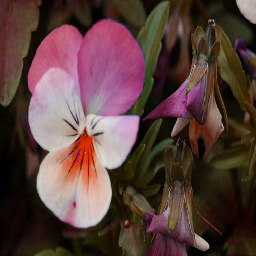}
 \includegraphics[width=0.17\linewidth]{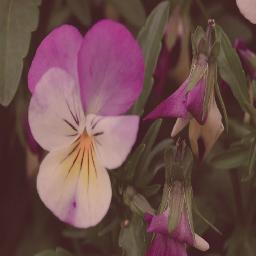} \\
\end{tabular}
\caption{Visual color transfer results of the Hue-Net compared to the results obtained by~\cite{reinhard2001color} and to the Hue-Net trained without the semantic-based (MI) loss $\pazocal{L}_{\scriptsize\mbox{MI}}.$ Examples are from the Oxford 102 Category Flower dataset~\cite{Nilsback08}.
The semantic-based loss encourages meaningful colorization - see for example the insect on the flower in the third row, the background flower in the fourth row, the stigma and the anther in the fifth row - all are marked with red rectangles.
\label{fig:TableOfFigures}}
\end{figure*}
\vspace{-.1cm}
\section{Experimental Results}\label{sec:ExperimentalResults}
\vspace{-.1cm}
We test our method on three datasets:
\begin{enumerate}
\item {\bf Oxford 102 Category Flower Dataset}~\cite{Nilsback08}
consists of 8189 images, which we divide into $7370$ and $819$ images for train and test respectively.
The images have large scale, pose and light variations. In addition, there are categories that have large variations within the category and several very similar categories. Example images are shown in Figure~\ref{fig:TableOfFigures}.
\item {\bf Flower Color Images} consists of only 210 images $(128\times 128\times 3)$ was downloaded from Kaggle web site: \url{https://www.kaggle.com/olgabelitskaya/flower-color-images}. Examples are shown in Figures~\ref{fig:CT-Hue},\ref{fig:CT-main} and in the Hue-Net architecture diagram (Figure~\ref{fig:Architecture}).
\item {\bf The Cars dataset} contains $16,185$ images of cars classified into $196$ classes~\cite{KrauseStarkDengFei-Fei_3DRR2013}. Examples are shown in Figure~\ref{fig:CT-Hue}.
\end{enumerate}


Evaluating the quality of synthesized images is an open
and difficult problem~\cite{salimans2016improved}. Traditional metrics such as per-pixel mean-squared error do not assess joint statistics of the result, and therefore do not measure the image similarity we aim to capture. A “real vs. fake” questionnaire based on our generated images and the true ones can be accessed via~\url{https://forms.gle/wAmRSChHpa65d1So7}. Human observer results and statistics based this questionnaire can be found in the supplementary material.

We evaluate the contribution of semantic-based Hue-Net loss by training the network without it. In addition, we compared the results to those obtained by Reinhard~\cite{reinhard2001color} via manipulations of the color histograms.
Results are shown in Figure~\ref{fig:TableOfFigures}.
As can be seen, the Hue-Net painting is not an application of deterministic functions to hue-channel histograms. The insect for example, that camouflages with the blue flower in the third row, appears to be green in the Hue-Net painted image while the petals are pink.   
Another example is the painted three flowers in Figure~\ref{fig:CT-main} (first row) that appear in different shades, as in the target image although they all have the same shade of purple in the source image.
For each of the images in Figure~\ref{fig:TableOfFigures}, we calculated the EMD and the relative MI between the target's and source's or output's hue channels. We also measured the relative MI for the color transfer results of the network trained without the MI loss.The results are shown in Table~\ref{tbl:EMDandMI}.
\begin{table}
\small
\begin{tabular}{ |c|c|c|c|c|c| }
 \hline
 & \multicolumn{2}{|c|}{\scriptsize{EMD}}
 & \multicolumn{3}{|c|}{\scriptsize{MI}}
\\
 \hline
 \scriptsize{No.}
& \scriptsize{TAR-SRC}
& \scriptsize{TAR-OUT}
& \scriptsize{SRC-TAR}
& \scriptsize{SRC-OUT}
& \scriptsize{SRC-OUT}
\\
& & & & & \scriptsize{w/o $\pazocal{L}_{\scriptsize\mbox{MI}}$}
\\
\hline
1 & .175 & .08  & .029 & .21 & .111\\
2 & .184 & .131 & .039 & .46 & .112\\
3 & .173 & .139 & .069 & .255 & .118\\
4 & .186 & .053 & .055 & .304 & .094\\
5 & .482 & .081 & .034 & .144 & .046\\
6 & .221 & .088 & .07 & .263 & .127\\
7 &	.29  & .216 & .036 & .18 & .095\\
\hline
\end{tabular}\caption{EMD between the hue-channel histograms of the output/source and the target. As expected (and desired) the intensity-based distance (cyclic EMD) between the output and the target is lower. 
Relative mutual information between the hue channels of the target/output and the source. We also calculated the mutual information for the output of a network trained without MI loss. As expected (and desired), the MI between the output and the source is higher using the complete Hue-Net loss.
}\label{tbl:EMDandMI}
\end{table}
\vspace{-.1cm}
\section{Summary and future work}
\vspace{-.1cm}
We presented the Hue-Net, a novel deep learning method for color transfer based on the construction of differentiable histograms and histogram-based loss functions. Specifically, intensity-based and semantic-based metrics are used to encourage intensity similarity to the target image and semantic similarity to the source image.
The adversarial loss is incorporated to constrain the generation of realistic images, making sure, for example, that the leaves and nor the petals will be painted in green.
While the results are promising we believe that the tools we developed can be applicable to different image-to-image translation tasks, such as photo enhancement, removal of illumination effects and colorization, with slight modifications.
{\small
\bibliographystyle{ieee_fullname}
\bibliography{egbib}
}

\end{document}